\documentclass[runningheads]{llncs}
\usepackage{graphicx}
\usepackage{mathtools}
\usepackage{amsmath}
\usepackage{array,varwidth}
\usepackage[para,online,flushleft]{threeparttable}
\usepackage{multirow}
\usepackage{comment}
\usepackage{verbatim}

\begin{document}
\title{Multi-objective point cloud autoencoders for explainable myocardial infarction prediction}
\titlerunning{Multi-objective point cloud autoencoders for explainable MI prediction}
%

%

\author{Marcel Beetz\inst{1} \and Abhirup Banerjee\inst{1,2}\orcidID{0000-0001-8198-5128} \and Vicente Grau\inst{1}\orcidID{0000-0001-8139-3480}}


\authorrunning{M. Beetz et al.}

%


\institute{Institute of Biomedical Engineering, Department of Engineering Science, University of Oxford, Oxford OX3 7DQ, United Kingdom \and 
Division of Cardiovascular Medicine, Radcliffe Department of Medicine, University of Oxford, Oxford OX3 9DU, United Kingdom
}

%
\maketitle              
\begin{abstract}

Myocardial infarction (MI) is one of the most common causes of death in the world. Image-based biomarkers commonly used in the clinic, such as ejection fraction, fail to capture more complex patterns in the heart's 3D anatomy and thus limit diagnostic accuracy. In this work, we present the multi-objective point cloud autoencoder as a novel geometric deep learning approach for explainable infarction prediction, based on multi-class 3D point cloud representations of cardiac anatomy and function. Its architecture consists of multiple task-specific branches connected by a low-dimensional latent space to allow for effective multi-objective learning of both reconstruction and MI prediction, while capturing pathology-specific 3D shape information in an interpretable latent space. Furthermore, its hierarchical branch design with point cloud-based deep learning operations enables efficient multi-scale feature learning directly on high-resolution anatomy point clouds. In our experiments on a large UK Biobank dataset, the multi-objective point cloud autoencoder is able to accurately reconstruct multi-temporal 3D shapes with Chamfer distances between predicted and input anatomies below the underlying images' pixel resolution. Our method outperforms multiple machine learning and deep learning benchmarks for the task of incident MI prediction by 19\% in terms of Area Under the Receiver Operating Characteristic curve. In addition, its task-specific compact latent space exhibits easily separable control and MI clusters with clinically plausible associations between subject encodings and corresponding 3D shapes, thus demonstrating the explainability of the prediction. 

\end{abstract}

\section{Introduction}
Myocardial infarction (MI) is the deadliest cardiovascular disease in the developed world \cite{khan2020global}. Consequently, an ability to predict future MI events is of immense importance on both an individual and population health level, as it would allow for improved risk stratification, preventative care, and treatment planning. In current clinical practice, MI prediction is typically based on volumetric biomarkers, such as ejection fraction. These can be derived from cardiac cine magnetic resonance imaging (MRI), which is considered the gold standard modality for cardiac function assessment \cite{reindl2020role}. While such metrics are relatively easy to calculate and interpret, they only approximate the complex 3D morphology and physiology of the heart with a single value, which hinders further improvements in predictive accuracy. Consequently, considerable research efforts have been dedicated to developing new methods capable of extracting novel biomarkers from images or segmentation masks using machine learning and deep learning techniques \cite{avard2022non,bernard2018deep,cetin2018radiomics,isensee2018automatic,khened2018densely,suinesiaputra2017statistical,wolterink2018automatic,zhang2019deep}. However, their focus on 2D data still limits the discovery of more intricate biomarkers whose important role for MI prediction and cardiac function assessment has previously been shown \cite{beetz2022interpretable,beetz2023mesh,corral2022understanding,suinesiaputra2017statistical}. In order to efficiently process true 3D anatomical shape information, geometric deep learning methods for point clouds have recently been increasingly used for various cardiac image-based tasks \cite{beetz2021generating,beetz2023point2mesh,beetz2021predicting,chang2020automatic,chen2021shape,ye2020pc,zhou2019one}.

In this work, we propose the multi-objective point cloud autoencoder as a novel geometric deep learning approach for interpretable MI prediction, based on 3D cardiac shape information. Its specialized multi-branch architecture allows for the direct and efficient processing of high resolution 3D point cloud representations of the multi-class cardiac anatomy at multiple time points of the cardiac cycle, while simultaneously predicting future MI events. Crucially, a low-dimensional latent space vector captures task-specific 3D shape information as an orderly multivariate probability distribution, offering pathology-specific separability and allowing for a straightforward visual analysis of associations between 3D structure and latent encodings. The resulting high explainability considerably boosts the method's clinical applicability and sets it apart from previous black-box deep learning approaches for MI classification \cite{beetz20233d,beetz2023post,chang2020automatic}. To the best of our knowledge, this is the first point cloud deep learning approach to combine full 3D shape processing and multi-objective learning with an explicit focus on method interpretability for MI prediction.

\section{Methods}

\subsection{Dataset and Preprocessing}
\label{ssec:dataset_and_preprocessing}
We select the cine MRI acquisitions of 470 subjects of the UK Biobank study as our dataset in this work \cite{petersen2013imaging}. All images were acquired with a voxel resolution of $1.8 \times 1.8 \times 8.0~\mbox{mm}^3$ for short-axis and $1.8 \times 1.8 \times 6.0~\mbox{mm}^3$ for long-axis slices using a balanced steady-state free precession (bSSFP) protocol \cite{petersen2015uk}. Half of the subjects in our dataset experienced an MI event after the image acquisition date (incident MI) as indicated by UK Biobank field IDs 42001 and 42000. The other 50\% of subjects are considered as normal control cases. They were chosen to be free of any cardiovascular disease and other pathologies frequently observed in the UK Biobank study, following a similar selection as previous works \cite{bai2020population,beetz2022multi,beetz2022combined} (see Table~\ref{tab:ukbb_codes_disease_definition}). 
For each subject, we reconstruct 3D multi-class point cloud representations of their biventricular anatomy from the corresponding raw cine MR images at both the end-diastolic (ED) and end-systolic (ES) phases of the cardiac cycle with the fully automatic multi-step process proposed in \cite{banerjee2021ptrsa,beetz2021biventricular,beetz2023multi}, and use them as inputs for our networks.

\subsection{Network Architecture}
\label{ssec:network_design}
\begin{figure}[t]
\centering
\includegraphics[width=1.0\textwidth]{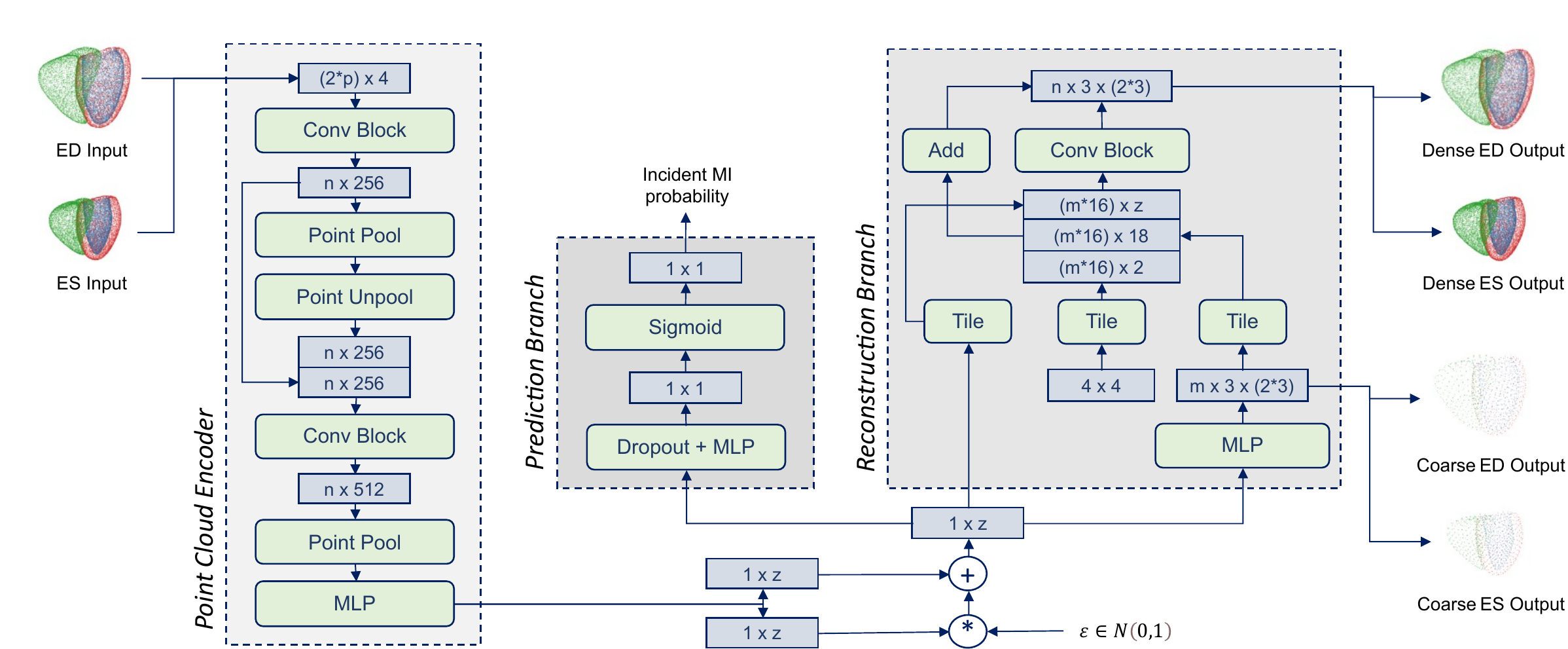}
\caption{Network architecture of the proposed multi-objective point cloud autoencoder. First, a point cloud deep learning-based encoder branch captures multi-scale shape information from multi-class and multi-temporal input anatomies in a low-dimensional latent space vector. Then, the resulting encodings are used in a reconstruction branch to recreate the original input shapes and in a prediction branch to output a clinical outcome probability (in this case for incident MI events).}
\label{fig:network_architecture}
\end{figure}

The architecture of the multi-objective point cloud autoencoder consists of three task-specific branches, namely an encoder, a reconstruction, and a prediction branch, which are connected by a low-dimensional latent space vector (Fig.~\ref{fig:network_architecture}).

Concatenated multi-class point clouds at the ED and ES phases of the cardiac cycle with shape $(2*p) \times 4$ are first fed into the encoder branch as network inputs where $(2*p)$ represents the number of points $p$ in the ED and ES point clouds and $4$ are the x, y, z coordinate values in 3D space and a class label to encode the three cardiac substructures, namely left ventricular (LV) endocardium, LV epicardium, and right ventricular (RV) endocardium. The inputs are then passed through the point cloud-specific encoder, which is composed of two connected PointNet-style \cite{qi2017pointnet,qi2017pointnet++} blocks and a multi-layer perceptron (MLP), before outputting both a mean and standard deviation (SD) vector of size $1 \times z$. 
Next, the reparameterization trick is applied to these two vectors, and the resulting latent space vector is used as an input to both the reconstruction and prediction branches. This ensures that the latent space is influenced by both tasks during training and thus encourages an interpretable distribution that is both discriminative enough for the prediction task and also descriptive enough to allow accurate reconstruction. 
The reconstruction branch \cite{yuan2018pcn} starts with a MLP to produce an intermediate coarse point cloud output, which assures that the final fine point cloud preserves the global shape. It is then followed by a FoldingNet-style \cite{yang2017foldingnet} layer to obtain the final dense output point cloud with both a local and global shape focus. The preliminary coarse and the dense output point cloud are represented as $m \times 3 \times (2*3)$ and $n \times 3 \times (2*3)$ tensors respectively, where $m$ and $n$ refer to the number of points with $n >> m$, the $3$ to the spatial 3D coordinates, and the $(2*3)$ to the three cardiac substructures at ED and ES. In this work, we use the same total number of points to represent both the input and dense output point clouds.
The prediction branch combines a Dropout layer, a MLP, and a Sigmoid activation function.

\subsection{Loss and Training}
\label{ssec:implementation_and_training}

The loss function of the multi-objective point cloud autoencoder consists of the sum of three subloss terms, each representing a different training objective in the multi-task setting, and weighted by two parameters $\beta$ and $\gamma$.

\begin{equation}
L_{total} = L_{reconstruction} + \beta * L_{KL} + \gamma * L_{CE}.
\label{eq:total_loss}
\end{equation}

The first loss term, $L_{reconstruction}$, encourages the network to accurately reconstruct input anatomies and thereby capture important shape information. It contains two subloss terms and a weighting parameter $\alpha$.

\begin{equation}
L_{reconstruction} = \displaystyle\sum_{i = 1}^{T} \displaystyle\sum_{j = 1}^{C} \big(L_{coarse, i, j} + \alpha * L_{dense, i, j}\big).
\label{eq_loss}
\end{equation}
Here, $C$ and $T$ refer to the number of cardiac substructures and phases respectively. We use $C=3$ and $T=2$ in this work. The $L_{coarse}$ and $L_{dense}$ loss terms compare the respective coarse and dense output predictions of the network with the same input point cloud using the symmetric Chamfer distance (CD). The weighting parameter $\alpha$ is increased stepwise from smaller (0.01) to larger (2.0) values during training in a monotonic annealing schedule to encourage the network to first focus on a good global reconstruction and gradually put more emphasis on a high local accuracy as training progresses.
The second loss term in Eq.~(\ref{eq:total_loss}), $L_{KL}$, calculates the Kullback-Leibler divergence between the network's latent space and a multivariate standard normal distribution, which encourages high latent space quality and improves regularization. The third loss term, $L_{CE}$, refers to the binary cross entropy loss between the network's outcome prediction and the gold standard encoding. We again use a monotonic annealing schedule for the weighting parameter $\beta$ to balance latent space quality and output accuracy and for $\gamma$ to gradually put more focus on improving prediction performance. Hereby, we choose stepwise increases from 0.001 to 0.01 for $\beta$ and from 1.0 to 5.0 for $\gamma$, based on empirical findings.

We randomly split the dataset into 70\% training, 5\% validation, and 25\% test data. We train the network with the Adam optimizer and a mini-batch size of 8 for $\sim$80,000 steps, since no improvement on the validation data was achieved during the 10,000 prior steps. The method is implemented using the TensorFlow library and has a post-training run time of $\sim$15~ms. All experiments are performed on a GeForce RTX 2070 Graphics Card with 8~GB memory.

\section{Experiments and Results}

\subsection{Input Shape Reconstruction}
\label{ssec:prediction_quality}

\begin{figure}[t]
\centering
\includegraphics[width=1.0\textwidth]{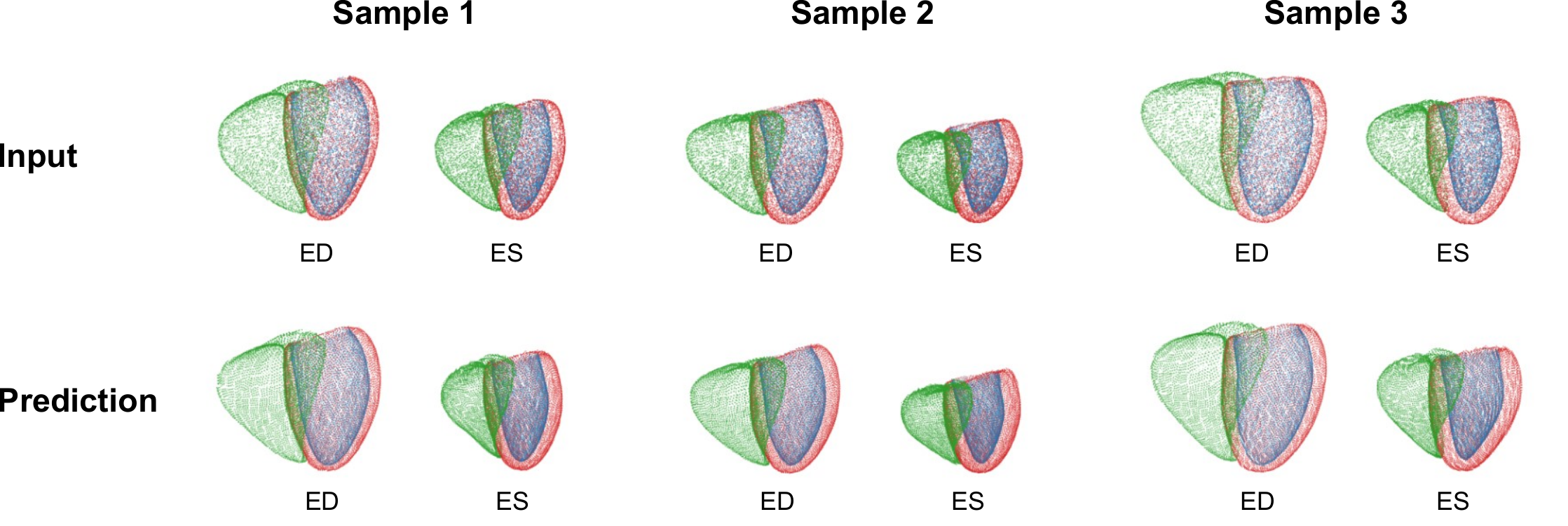}
\caption{Qualitative reconstruction results of three sample cases.}
\label{fig:qual_recon_results}
\end{figure}

In our first experiment, we evaluate whether the multi-objective point cloud autoencoder is able to accurately reconstruct the ED and ES input anatomies. To this end, we pass all anatomies of the test dataset through the trained network and visualize both the input and corresponding predicted point clouds of three sample cases in Fig.~\ref{fig:qual_recon_results}. We observe good local and global shape alignment between the input and predicted anatomies in all cases. Relationships between cardiac substructures and between ED and ES phases are accurately retained.

Next, we quantify the reconstruction performance by calculating the symmetric Chamfer distances between the respective input and reconstructed point clouds of all subjects in the test dataset separately for each cardiac substructure and phase (Table~\ref{tab:recon_results_chamfer}). We find mean Chamfer distance values below the underlying acquisition's pixel resolution for both phases and all cardiac substructures.

\begin{table*}[!h]
    \caption{Reconstruction results of the proposed method.}
    \def\arraystretch{1.5}\tabcolsep=3pt
    \centering
    \begin{tabular}{p{50pt}p{30pt}p{72pt}p{72pt}p{72pt}}
        \hline
        Metric & Phase & LV endocardium & LV epicardium & RV endocardium \\
        \hline
        \multirow{2}{*}{CD (mm)} & ED   & 1.57 ($\pm$0.35) & 1.53 ($\pm$0.23) & 1.71 ($\pm$0.27) \\
         & ES   & 1.26 ($\pm$0.30) & 1.47 ($\pm$0.29) & 1.67 ($\pm$0.32)\\                
        \hline
        \multicolumn{5}{@{}l}{Values represent mean ($\pm$SD). CD = Chamfer distance.}
    \end{tabular}
    \label{tab:recon_results_chamfer}
\end{table*}

\subsection{Myocardial Infarction Prediction}
\label{ssec:clinical_metrics}

We next evaluate the performance of the network for incident MI prediction as its second task. To this end, we first obtain both the gold standard MI outcomes and the MI predictions of our pre-trained network for all cases in the test dataset and quantify its performance using five common binary classification metrics (Table~\ref{tab:mi_pred_results}). To compare with clinical benchmarks, we select LV ejection fraction (EF) and the combination of LV and RV EF as widely used metrics and use each of them as input features for two separate logistic regression models. In addition, we choose a hierarchical convolutional neural network (CNN) and a standard PointNet \cite{qi2017pointnet} with 2D segmentation masks and 3D anatomy point clouds at ED and ES as respective inputs, as additional benchmarks (Table~\ref{tab:mi_pred_results}).

\begin{table*}[!h]
    \caption{Comparison of MI prediction results by multiple methods.}
    \def\arraystretch{1.5}\tabcolsep=3pt
    \centering
    \begin{tabular}{p{55pt}p{60pt}p{38pt}p{38pt}p{38pt}p{30pt}p{40pt}}
        \hline
        Input & Method & AUROC & Accuracy & Precision & Recall & F1-Score  \\
        \hline
        LV EF  &  Regression  &  0.622  &  0.570 &  0.591 &  0.504 &  0.533 \\
        LV+RV EF  &  Regression  &  0.611  &  0.571 &  0.499 &  0.516 &  0.540 \\
        2D shapes  &  CNN  & 0.641  &  0.608 &  0.603 &  0.633 &  0.617 \\
        3D shape  &  PointNet  & 0.646  &  0.652 &  0.666 &  0.610 &  0.637 \\
        3D shape  &  Proposed  & \textbf{0.767}  &  \textbf{0.694} &  \textbf{0.706} &  \textbf{0.683} &  \textbf{0.695} \\
        \hline
    \end{tabular}
    \label{tab:mi_pred_results}
\end{table*}

We find that the proposed multi-objective point cloud autoencoder outperforms all other approaches with improvements of 19\% in terms of Area Under the Receiver Operating Characteristic (AUROC) curve.

\subsection{Task-Specific Latent Space Analysis}
\begin{figure}[t]
\centering
\includegraphics[width=1.0\textwidth]{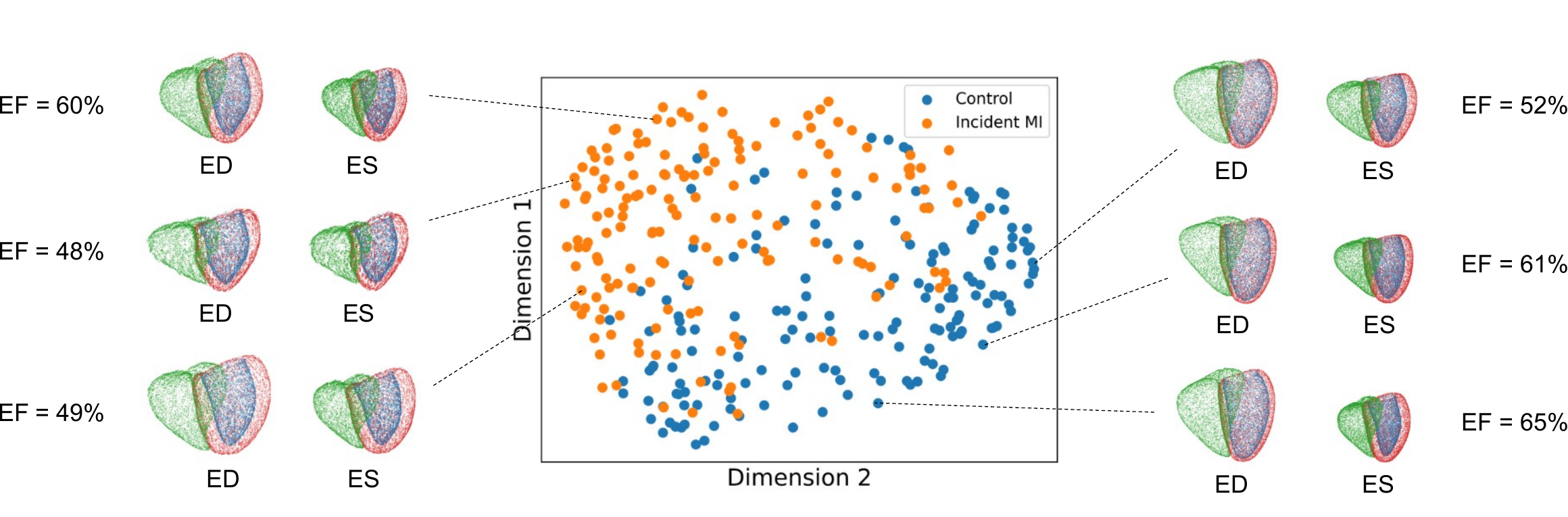}
\caption{Laplacian eigenmap of latent space encodings of both normal (blue) and MI (orange) subjects. ED and ES anatomies with LV ejection fraction (EF) values are shown for 6 cases located in salient map regions.}
\label{fig:laplacian_eigenmap}
\end{figure}

In addition to validating the reconstruction and prediction performance of our network, we also investigate the ability of its latent space to store high-resolution 3D shape data in an interpretable and pathology-specific manner. To this end, we first pass the anatomy point clouds of both normal and MI cases through the encoder branch of the pre-trained network to obtain their respective latent space encodings. We then apply the Laplacian eigenmap \cite{belkin2001laplacian} algorithm to the encodings as a non-linear dimensionality reduction technique and visualize the resulting 2D eigenmap of the latent space distribution in Fig.~\ref{fig:laplacian_eigenmap}. In addition, in order to study associations between the latent subject encodings and their 3D anatomical shapes, we select 6 cases encoded at salient locations in the eigenmap and plot their pertinent 3D anatomies (Fig.~\ref{fig:laplacian_eigenmap}).

We observe a clear differentiation between the encoded normal and MI cases in the eigenmap. Furthermore, the sample anatomies positioned in the area of normal subjects typically exhibit noticeably different shape patterns to the ones located in the MI cluster. For example, the left middle MI subject shows much smaller volume and myocardial thickness changes between ED and ES anatomies than the three normal cases.

\subsection{Ablation Study}
\label{ssec:pathology_deformation}
In order to assess the contributions of important parts of our network architecture, we next ablate multiple key components and study their effects on prediction performance. More specifically, we individually remove the dropout layer, the KL loss term, and the reconstruction branch, retrain each of the three ablated networks, and report their MI prediction results on the test dataset in Table~\ref{tab:ablation_results}. In addition, we investigate the importance of the multi-objective setting by first training the point cloud autoencoder without a prediction branch with a single reconstruction objective and then applying a logistic regression model for MI prediction to the learned general-purpose latent space representation (Table~\ref{tab:ablation_results}).

\begin{table*}[!h]
    \caption{Effects of architecture ablations on prediction performance.}
    \def\arraystretch{1.5}\tabcolsep=3pt
    \centering
    \begin{tabular}{p{120pt}p{38pt}p{38pt}p{38pt}p{30pt}p{40pt}}
        \hline
        Method & AUROC & Accuracy & Precision & Recall & F1-Score  \\
        \hline
        Proposed  & \textbf{0.767}  &  \textbf{0.694} &  0.706 &  \textbf{0.683} &  \textbf{0.695} \\
        W/o dropout  & 0.739  &  0.652 &  0.655 &  0.667 &  0.661 \\
        W/o KL loss term & 0.731  &  0.678 & 0.689 &  0.667 &  0.677 \\
        W/o reconstruction branch  & 0.755  &  0.686 &  \textbf{0.744} &  0.583 &  0.654 \\
        W/o multi-objective training  & 0.717  &  0.655 &  0.659 &  0.648 &  0.648 \\
        \hline
    \end{tabular}
    \label{tab:ablation_results}
\end{table*}

All components contributed positively to the overall prediction performance with multi-objective training having the largest effect. The network without a reconstruction branch achieved the second-best AUROC, highest precision, and lowest recall score.

\section{Discussion and Conclusion}
In this paper, we have presented the multi-objective point cloud autoencoder as a novel geometric deep learning approach for interpretable MI prediction. The network is able to reconstruct input point clouds with high accuracy and only small localized smoothness artifacts despite the difficult multi-task setting. This shows the suitability of its architecture for efficient multi-scale feature extraction and its ability to effectively capture important 3D shape information in its latent space. Furthermore, the network can simultaneously process all three cardiac substructures at both ED and ES, indicating high flexibility and a potential for further extensions to the full cardiac cycle or other cardiac substructures. In addition, it also allows for more complex 3D shape-based biomarkers to be learned based on inter-temporal and inter-anatomical relationships. All these results are achieved directly on point cloud data, which offers a considerably more efficient storage of anatomical surface information than widely used voxelgrid-based deep learning approaches. Furthermore, the method is fast, fully automatic, and can be readily incorporated into a 3D shape analysis pipeline with cine MRI inputs.

The network also outperforms both machine learning techniques based on widely used clinical biomarkers as well as other deep learning approaches for MI prediction. On the one hand, this corroborates previous findings on the increased utility of full 3D shape information compared to single-valued or 2D biomarkers for MI assessment \cite{beetz2023post,corral2022understanding,suinesiaputra2017statistical}. On the other hand, it shows the higher capacity of the proposed architecture and training process to extract important novel 3D biomarkers relevant for MI prediction. While we only study MI classification as a sample use case in this work, we believe that the proposed approach can be easily applied to other 3D shape-related pathologies or risk factors.

The network achieves these results based on a highly interpretable latent space with a clear differentiation between normal and MI subject encodings. Furthermore, the observed associations between encodings and 3D shapes demonstrate that the latent space is not only discriminative but also that the differentiation is based on clinically plausible 3D shape differences, such as reduced myocardial thinning between ED and ES in MI subjects which is indicative of impaired contraction ability of the heart. This greatly improves the explainability and applicability of the approach, as new subject phenotypes can be quickly and easily compared to other ones with similar encodings. Furthermore, the latent map not only shows well known associations of EF and MI but also a clear differentiation between some normal and MI cases with similar EF values. This indicates that the network is able to capture more intricate biomarkers that go beyond ejection fraction and to successfully utilize them in its MI prediction task while retaining high interpretability.

Finally, we show in our ablation studies that all major components of the architecture improve predictive accuracy. We hypothesize that the dropout layer, KL divergence term, and reconstruction branch introduce useful constraints, which have a positive regularizing effect and aid generalization. The multi-objective training procedure accounts for the largest performance gain. This is likely due to the exploited synergies of multiple tasks, which we also believe to be the primary reason for the high separability in the latent space.

\section*{Acknowledgments}
This research has been conducted using the UK Biobank Resource under Application Number ‘40161’. The authors express no conflict of interest. The work of M. Beetz was supported by the Stiftung der Deutschen Wirtschaft (Foundation of German Business). A. Banerjee is a Royal Society University Research Fellow and is supported by the Royal Society Grant No. URF{\textbackslash}R1{\textbackslash}221314. The work of A. Banerjee was partially supported by the British Heart Foundation (BHF) Project under Grant PG/20/21/35082. The work of V. Grau was supported by the CompBioMed 2 Centre of Excellence in Computational Biomedicine (European Commission Horizon 2020 research and innovation programme, grant agreement No. 823712).

\begin{table*}[htbp]
    \caption{Pathologies with corresponding codes of UK Biobank field ID 20002 used to select control cases in this work.}
    \def\arraystretch{1.5}\tabcolsep=3pt
    \centering
    \begin{tabular}{|p{20pt}p{144pt}|p{20pt}p{140pt}|}
        \hline
        Code & Meaning & Code & Meaning\\
        \hline
        1065 & Hypertension & 1286 & Depression \\
        1066 & Heart/cardiac problem & 1412 & Bronchitis \\
        1067 & Peripheral vascular disease & 1471 & Atrial fibrillation \\
        1072 & Essential hypertension & 1472 & Emphysema \\
        1073 & Gestational hypertension/pre-eclampsia & 1473 & High cholesterol \\
        1074 & Angina & 1483 & Atrial flutter \\
        1075 & Heart attack/myocardial infarction & 1484 & Wolff Parkinson white/WPW syndrome \\
        1076 & Heart failure/pulmonary odema & 1485 & Irregular heart beat \\
        1077 & Heart arrhythmia & 1486 & Sick sinus syndrome \\
        1078 & Heart valve problem/heart murmur & 1487 & SVT/supraventricular tachycardia \\
        1079 & Cardiomyopathy & 1491 & Brain haemorrhage \\
        1080 & Pericardial problem & 1492 & Aortic aneurysm \\
        1081 & Stroke & 1496 & Alpha-1 antitrypsin deficiency \\
        1086 & Subarachnoid haemorrhage & 1531 & Post-natal depression \\
        1087 & Leg claudication/intermittent claudication & 1583 & Ischaemic stroke \\
        1088 & Arterial embolism & 1584 & Mitral valve disease \\
        1111 & Asthma & 1585 & Mitral regurgitation/incompetence \\
        1112 & Chronic obstructive airways disease/COPD & 1586 & Aortic valve disease \\
        1113 & Emphysema/chronic bronchitis & 1587 & Aortic regurgitation/incompetence \\
        1220 & Diabetes & 1588 & Hypertrophic cardiomyopathy \\
        1221 & Gestational diabetes & 1589 & Pericarditis \\
        1222 & Type 1 diabetes & 1590 & Pericardial effusion \\
        1223 & Type 2 diabetes & 1591 & Aortic aneurysm rupture \\
        1262 & Parkinson’s disease & 1592 & Aortic dissection \\
        1263 & Dementia/Alzheimer’s/cognitive impairment & & \\
        \hline
        \multicolumn{4}{@{}l}{}
    \end{tabular}
    \label{tab:ukbb_codes_disease_definition}
\end{table*}

%
%
%
\bibliographystyle{splncs04}
\bibliography{refs.bib}

\begin{thebibliography}{10}
\providecommand{\url}[1]{\texttt{#1}}
\providecommand{\urlprefix}{URL }
\providecommand{\doi}[1]{https://doi.org/#1}

\bibitem{avard2022non}
Avard, E., et~al.: Non-contrast cine cardiac magnetic resonance image radiomics
  features and machine learning algorithms for myocardial infarction detection.
  Computers in Biology and Medicine  \textbf{141},  105145 (2022)

\bibitem{bai2020population}
Bai, W., et~al.: A population-based phenome-wide association study of cardiac
  and aortic structure and function. Nature Medicine  \textbf{26}(10),
  1654--1662 (2020)

\bibitem{banerjee2021ptrsa}
Banerjee, A., et~al.: A completely automated pipeline for {3D} reconstruction
  of human heart from {2D} cine magnetic resonance slices. Philosophical
  Transactions of the Royal Society A: Mathematical, Physical and Engineering
  Sciences  \textbf{379}(2212),  20200257 (2021)

\bibitem{beetz2021biventricular}
Beetz, M., Banerjee, A., Grau, V.: Biventricular surface reconstruction from
  cine {MRI} contours using point completion networks. In: 2021 IEEE 18th
  International Symposium on Biomedical Imaging (ISBI). pp. 105--109. IEEE
  (2021)

\bibitem{beetz2021generating}
Beetz, M., Banerjee, A., Grau, V.: Generating subpopulation-specific
  biventricular anatomy models using conditional point cloud variational
  autoencoders. In: International Workshop on Statistical Atlases and
  Computational Models of the Heart. pp. 75--83. Springer (2021)

\bibitem{beetz2022multi}
Beetz, M., Banerjee, A., Grau, V.: Multi-domain variational autoencoders for
  combined modeling of {MRI}-based biventricular anatomy and {ECG}-based
  cardiac electrophysiology. Frontiers in Physiology p.~991 (2022)

\bibitem{beetz2023point2mesh}
Beetz, M., Banerjee, A., Grau, V.: {Point2Mesh-Net}: Combining point cloud and
  mesh-based deep learning for cardiac shape reconstruction. In: International
  Workshop on Statistical Atlases and Computational Models of the Heart. pp.
  280--290. Springer (2023)

\bibitem{beetz2021predicting}
Beetz, M., et~al.: Predicting {3D} cardiac deformations with point cloud
  autoencoders. In: International Workshop on Statistical Atlases and
  Computational Models of the Heart. pp. 219--228. Springer (2021)

\bibitem{beetz2022combined}
Beetz, M., et~al.: Combined generation of electrocardiogram and cardiac anatomy
  models using multi-modal variational autoencoders. In: 2022 IEEE 19th
  International Symposium on Biomedical Imaging (ISBI). pp.~1--4 (2022)

\bibitem{beetz2022interpretable}
Beetz, M., et~al.: Interpretable cardiac anatomy modeling using variational
  mesh autoencoders. Frontiers in Cardiovascular Medicine p.~3258 (2022)

\bibitem{beetz20233d}
Beetz, M., et~al.: {3D} shape-based myocardial infarction prediction using
  point cloud classification networks. arXiv preprint arXiv:2307.07298  (2023)

\bibitem{beetz2023mesh}
Beetz, M., et~al.: Mesh {U-Nets} for {3D} cardiac deformation modeling. In:
  International Workshop on Statistical Atlases and Computational Models of the
  Heart. pp. 245--257. Springer (2023)

\bibitem{beetz2023multi}
Beetz, M., et~al.: Multi-class point cloud completion networks for {3D} cardiac
  anatomy reconstruction from cine magnetic resonance images. arXiv preprint
  arXiv:2307.08535  (2023)

\bibitem{beetz2023post}
Beetz, M., et~al.: Post-infarction risk prediction with mesh classification
  networks. In: International Workshop on Statistical Atlases and Computational
  Models of the Heart. pp. 291--301. Springer (2023)

\bibitem{belkin2001laplacian}
Belkin, M., Niyogi, P.: Laplacian eigenmaps and spectral techniques for
  embedding and clustering. Advances in Neural Information Processing Systems
  \textbf{14} (2001)

\bibitem{bernard2018deep}
Bernard, O., et~al.: Deep learning techniques for automatic {MRI} cardiac
  multi-structures segmentation and diagnosis: is the problem solved? IEEE
  Transactions on Medical Imaging  \textbf{37}(11),  2514--2525 (2018)

\bibitem{cetin2018radiomics}
Cetin, I., et~al.: A radiomics approach to computer-aided diagnosis with
  cardiac cine-{MRI}. In: International Workshop on Statistical Atlases and
  Computational Models of the Heart. pp. 82--90. Springer (2018)

\bibitem{chang2020automatic}
Chang, Y., Jung, C.: Automatic cardiac {MRI} segmentation and
  permutation-invariant pathology classification using deep neural networks and
  point clouds. Neurocomputing  \textbf{418},  270--279 (2020)

\bibitem{chen2021shape}
Chen, X., et~al.: Shape registration with learned deformations for {3D} shape
  reconstruction from sparse and incomplete point clouds. Medical Image
  Analysis  \textbf{74},  102228 (2021)

\bibitem{corral2022understanding}
Corral~Acero, J., et~al.: Understanding and improving risk assessment after
  myocardial infarction using automated left ventricular shape analysis. JACC:
  Cardiovascular Imaging  (2022)

\bibitem{isensee2018automatic}
Isensee, F., et~al.: Automatic cardiac disease assessment on cine-{MRI} via
  time-series segmentation and domain specific features. In: International
  Workshop on Statistical Atlases and Computational Models of the Heart. pp.
  120--129. Springer (2018)

\bibitem{khan2020global}
Khan, M.A., et~al.: Global epidemiology of ischemic heart disease: results from
  the global burden of disease study. Cureus  \textbf{12}(7) (2020)

\bibitem{khened2018densely}
Khened, M., et~al.: Densely connected fully convolutional network for
  short-axis cardiac cine {MR} image segmentation and heart diagnosis using
  random forest. In: International Workshop on Statistical Atlases and
  Computational Models of the Heart. pp. 140--151. Springer (2018)

\bibitem{petersen2013imaging}
Petersen, S.E., et~al.: Imaging in population science: cardiovascular magnetic
  resonance in 100,000 participants of {UK Biobank} - rationale, challenges and
  approaches. Journal of Cardiovascular Magnetic Resonance  \textbf{15}(46),
  1--10 (2013)

\bibitem{petersen2015uk}
Petersen, S.E., et~al.: {UK Biobank’s} cardiovascular magnetic resonance
  protocol. Journal of cardiovascular magnetic resonance  \textbf{18}(8), ~1--7
  (2016)

\bibitem{qi2017pointnet}
Qi, C.R., et~al.: Pointnet: Deep learning on point sets for {3D} classification
  and segmentation. In: Proceedings of the IEEE Conference on Computer Vision
  and Pattern Recognition. pp. 652--660 (2017)

\bibitem{qi2017pointnet++}
Qi, C.R., et~al.: Pointnet++: Deep hierarchical feature learning on point sets
  in a metric space. In: Advances in Neural Information Processing Systems. pp.
  5099--5108 (2017)

\bibitem{reindl2020role}
Reindl, M., et~al.: Role of cardiac magnetic resonance to improve risk
  prediction following acute {ST}-elevation myocardial infarction. Journal of
  Clinical Medicine  \textbf{9}(4), ~1041 (2020)

\bibitem{suinesiaputra2017statistical}
Suinesiaputra, A., et~al.: Statistical shape modeling of the left ventricle:
  myocardial infarct classification challenge. IEEE Journal of Biomedical and
  Health Informatics  \textbf{22}(2),  503--515 (2017)

\bibitem{wolterink2018automatic}
Wolterink, J.M., et~al.: Automatic segmentation and disease classification
  using cardiac cine {MR} images. In: International Workshop on Statistical
  Atlases and Computational Models of the Heart. pp. 101--110. Springer (2018)

\bibitem{yang2017foldingnet}
Yang, Y., et~al.: Foldingnet: Interpretable unsupervised learning on {3D} point
  clouds. arXiv preprint arXiv:1712.07262  (2017)

\bibitem{ye2020pc}
Ye, M., et~al.: {PC-U} net: Learning to jointly reconstruct and segment the
  cardiac walls in {3D} from {CT} data. In: International Workshop on
  Statistical Atlases and Computational Models of the Heart. pp. 117--126.
  Springer (2020)

\bibitem{yuan2018pcn}
Yuan, W., et~al.: {PCN}: Point completion network. In: 2018 International
  Conference on 3D Vision (3DV). pp. 728--737 (2018)

\bibitem{zhang2019deep}
Zhang, N., et~al.: Deep learning for diagnosis of chronic myocardial infarction
  on nonenhanced cardiac cine {MRI}. Radiology  \textbf{291}(3),  606--617
  (2019)

\bibitem{zhou2019one}
Zhou, X.Y., et~al.: One-stage shape instantiation from a single {2D} image to
  {3D} point cloud. In: Medical Image Computing and Computer Assisted
  Intervention--MICCAI 2019: 22nd International Conference, Shenzhen, China,
  October 13--17, 2019, Proceedings, Part IV 22. pp. 30--38. Springer (2019)

\end{thebibliography}

\end{document}